\documentclass[sigconf]{aamas} 

\usepackage{balance} 
\usepackage{xcolor}

\usepackage{amsfonts}
\usepackage{tabularx}
\usepackage{graphics}
\usepackage{mdwlist}
\usepackage{xspace}
\usepackage{amsmath}
\usepackage{mathtools}
\usepackage{wrapfig}
\usepackage{makecell}
\newcommand{\hideContent}[1]{}
\usepackage{graphicx}
\usepackage{amsmath} 
\usepackage{booktabs}
\usepackage{soul}
\usepackage{multirow}
\usepackage{bigints}
\usepackage{amsthm}
\newtheorem{remark}{Remark}

\newcommand{\ourmethod}{\textsc{GradABM}\xspace}
\newcommand{\calibnn}{\textsc{CalibNN}\xspace}

\setcopyright{ifaamas}
\acmConference[AAMAS '23]{Proc.\@ of the 22nd International Conference
on Autonomous Agents and Multiagent Systems (AAMAS 2023)}{May 29 -- June 2, 2023}
{London, United Kingdom}{A.~Ricci, W.~Yeoh, N.~Agmon, B.~An (eds.)}
\copyrightyear{2023}
\acmYear{2023}
\acmDOI{}
\acmPrice{}
\acmISBN{}


\acmSubmissionID{165}

\title[Differentiable Agent-based Epidemiology]{Differentiable Agent-based Epidemiology}

\author{Ayush Chopra}
\authornote{Both authors contributed equally to this research.}
\affiliation{%
 \institution{Massachusetts Institute of Technology}
 \country{USA}
 }

\author{Alexander Rodr\'iguez}
\authornotemark[1]
\affiliation{
\institution{Georgia Institute of Technology}
\country{USA}
 }

 \author{Jayakumar Subramanian}
\affiliation{%
 \institution{Adobe MDSR}
 \country{India}
 }
 
 \author{Arnau Quera-Bofarull}
\affiliation{%
 \institution{University of Oxford}
 \country{United Kingdom}
 }

 \author{Balaji Krishnamurthy}
\affiliation{%
\institution{Adobe MDSR}
 \country{India}
 }

  \author{B. Aditya Prakash}
\affiliation{%
\institution{Georgia Institute of Technology}
 \country{USA}
 }

  \author{Ramesh Raskar}
\affiliation{%
\institution{Massachusetts Institute of Technology}
 \country{USA}
 }

\begin{abstract}
Mechanistic simulators are an indispensable tool for epidemiology to explore the behavior of complex, dynamic infections under varying conditions and navigate uncertain environments. Agent-based models (ABMs) are an increasingly popular simulation paradigm that can represent the heterogeneity of contact interactions with granular detail and agency of individual behavior. However, conventional ABM frameworks are not differentiable and present challenges in scalability; due to which it is non-trivial to connect them to auxiliary data sources. In this paper, we introduce \ourmethod: a scalable, differentiable design for agent-based modeling that is amenable to gradient-based learning with automatic differentiation. \ourmethod can quickly simulate million-size populations in few seconds on commodity hardware, integrate with deep neural networks and ingest heterogeneous data sources. This provides an array of practical benefits for calibration, forecasting, and evaluating policy interventions. We demonstrate the efficacy of \ourmethod via extensive experiments with real COVID-19 and influenza datasets.
\end{abstract}


\newcommand{\BibTeX}{\rm B\kern-.05em{\sc i\kern-.025em b}\kern-.08em\TeX}

\begin{document}


\pagestyle{fancy}
\fancyhead{}


\maketitle 


\section{Introduction}

Agent-based models (ABMs) are discrete simulators which comprise a collection of agents which can act and interact within a computational world~\cite{abm1, abm2, bio_abm, ai_economist, romero-brufau_public_2021}. They can explicitly represent heterogeneity of the interacting population via underlying contact networks and model the adaptability of individual behavior to enable more realistic simulations. As a result, ABMs are increasingly popular in epidemiology in place of classical ODE-based models, such as the SIR (susceptible-infected-recovered) model~\cite{hethcote2000mathematics,dimitrov2010mathematical}. In recent efforts to contain the spread of COVID-19, ABMs have been used to evaluate the benefit of delaying 2nd dose of the vaccine~\cite{romero-brufau_public_2021}, deploying mobile apps for digital contact tracing~\cite{abueg2021modeling}, and prioritizing test speed over specificity~\cite{larremore2021test}. The utility of such simulators for practical decision making depends upon their ability to recreate the population with great detail, integrate with real-world data streams, and analyze the sensitivity of results.

\begin{figure*}
    \centering
    \includegraphics[width=0.89\textwidth]{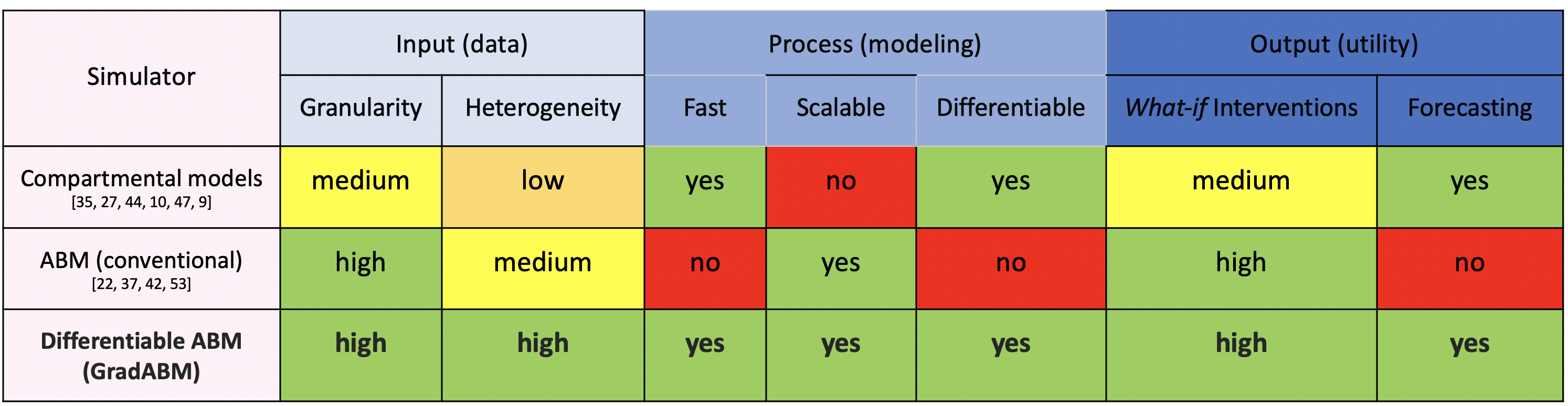}
    \caption{We introduce \ourmethod, an end-to-end differentiable ABM which can: i) ingest multi-granular (citizen, block, county, state) and heterogeneous (clinical, behavioral, policy, census, survey) data sources; ii) simulate realistic populations \textit{quickly and reliably} with gradient-based calibration; iii) facilitate flexible interventions for policy decisions and efficient forecasting with end-to-end DNN-\ourmethod pipelines. The key idea of \ourmethod is a general-purpose differentiable sparse tensor-calculus design which we validate by demonstrating utility for epidemiology here.}
    \label{fig:motivate}
\end{figure*}

However, ABMs are conventionally slow, difficult to scale to large population sizes~\cite{pellis2015eight} and tough to calibrate with real-world data~\cite{edeling2020model}. 
Conventional ABM frameworks follow an object-oriented (agent-centered) design where the agents are modeled as objects. While conceptually appealing, object-oriented approaches often represent inter-agent infection transmission (over contact networks) and within-agent disease progression inefficiently, resulting in poor scalability as the agent population and number of interactions grows. Performing a single forward simulation over a large ABM can take several days~\cite{alex-survey-43, aylett2021june}, making calibration results subpar where approaches require multiple forward passes~\cite{chang2021supporting,romero-brufau_public_2021,alex-survey-316}. This is an issue since simulation results (emergent behavior) can be highly sensitive to the scale of the input population and parameters. In addition, incorporating novel sources of data that could inform calibration and other downstream tasks (e.g., forecasting) is often laborious and adds overhead complexity to the ABM (e.g., incorporating digital exposure data to ABMs~\cite{hinch2021openabm}). 

In this paper, we introduce \ourmethod to alleviate these concerns and enhance the potential of ABMs for practical decision making in epidemiology (Fig.~\ref{fig:motivate}). Our key insight is to make simulations tensorized and differentiable which enables gradient-based learning with automatic differentiation. First, in contrast to conventional ABMs, \ourmethod follows a tensorized design~\cite{deepabm} where agent states are represented as vectors and their interaction networks as (sparse) adjacency matrices. As a result, forward simulation of agent interaction and disease progression can be run rapidly, in a highly parallelized manner (millions-size networked populations in a few seconds on commodity hardware). 
Second, ABMs are stochastic simulators which routinely require sampling from discrete distributions for execution (e.g., all interactions with an infected agent may not result in a new infection; COVID-19 tests may return false positives/negatives with some probability). Ensuring differentiability requires gradient estimation through such discrete stochasticity. To achieve this, \ourmethod reparameterizes discrete distributions (e.g., Bernoulli) with continuous relaxations using the gumbel-softmax gradient estimator~\cite{gumbel_softmax, gumbel_estimator_review}. This allows \ourmethod to benefit from gradient-based optimization and tools for automatic differentiation.

Using \ourmethod's scalable and differentiable design, we merge it with deep neural networks (DNNs) to help with calibration, forecasting and policy evaluation. In particular, we demonstrate how our method can be coupled with a calibration DNN allowing seamless integration of novel datasets from heterogeneous sources--several of which were difficult to incorporate in the calibration of previous ABM designs. 
To illustrate the effectiveness of our proposed framework, we run several experiments using real world datasets. In particular, we treat COVID-19 and Influenza as case studies, highlighting \ourmethod's benefits with respect to robust, time-space personalized forecasting, and evaluation of policy interventions.

\textbf{Our Contributions}: 
1) Present \ourmethod, which is a scalable and differentiable ABM design amenable for gradient-based learning with auto-diff.
2) Demonstrate that \ourmethod allows quickly simulating realistic population sizes, integrating with deep neural networks and ingesting heterogeneous data sources which provides an array of practical benefits for calibration, forecasting, and policy evaluation.
3) Validate effectiveness of \ourmethod through experiments using real-world datasets of COVID-19 and Influenza.

\section{Background and Related Work}
\label{sec:rw}
Before describing \ourmethod, we first review relevant works in agent-based models (ABMs) and with relevance to epidemiology. Then, we present our work in context of broad research in differentiable simulators and also briefly mention an insight we find useful when designing \ourmethod (permutation invariance). We refer the reader to~\cite{rodriguez2022data,dimitrov2010mathematical, hethcote2000mathematics, simulation-intelligence} for a more extensive discussion.

\subsection{Agent-based Modeling}
ABMs represent systems ~\cite{abm1,abm2} as collections of agents which can act and interact with each other within a computational world. Owing to their flexibility, practitioners have applied ABMs in a wide variety of problem domains such as modeling cells in a tumor micro-environment to diagnose cancers~\cite{bio_abm}, humans in a physical environment to study economic policies~\cite{ai_economist}, infectious diseases~\cite{romero-brufau_public_2021, abueg2021modeling} and avatars in a digital world to study misinformation~\cite{misinfo_abm}.

\vspace{2mm}
\noindent\textbf{ABMs in Epidemiology.} In the context of epidemiology, ABMs are used to understand how disease spreads and evaluate efficacy of health interventions~\cite{marathe2013computational}. They simulate transmission of pathogens within multiple contact networks with an epidemiological model~\cite{hinch2021openabm, aylett2021june} used to describe between-host transmission and within-host progression of infection. \cite{pellis2015eight, eames2015six} survey some methods and assumptions for designing such simulators. Our work introduces a tensorized and differentiable epidemiological simulator that can generalize across multiple infections and model specifications.

\vspace{2mm}
\noindent\textbf{Scalability of ABMs.} It is a key consideration since modeling granular population details is computationally expensive. Conventional frameworks like Mesa~\cite{mesa} follow an object-oriented design which while ease to use (Python-API) is prohibitively slow to scale (require few hours for a single iteration). There have been attempts to reduce this burden and simulate realistic scale with distributed HPC systems and GPU-optimized implementations. EpiFast~\cite{alex-survey-43} demonstrated simulating large contact networks in a few minutes on distributed systems. However, the required compute for such executions is expensive and not widely available. OpenABM~\cite{hinch2021openabm} presents an optimized C++/Cuda API to build an epidemiological ABM for fast execution on commodity hardware but it is non-trivial for epidemiological practitioners to extend, whilst not being amenable to gradient-based learning. Motivated by recent work in molecular dynamics~\cite{schoenholz2020jax}, \ourmethod is implemented using highly optimized sparse tensor APIs of auto-diff packages (e.g. PyTorch/JAX) to accelerate and scale simulations while preserving ease of use.

\vspace{2mm}
\noindent\textbf{Calibration of ABMs.} This involves identifying appropriate values for latent parameters and is essential to ensure reliability of results. Generally, a hybrid strategy is used where some parameters are sourced from control trials conducted offline by clinical experts and then others are calibrated in-silico by achieving a goodness-of-fit between emergent ABM output and real-world macro data (e.g. number of deaths). The conventional technique is to use grid search~\cite{chang2021supporting, romero-brufau_public_2021}, beam search~\cite{alex-survey-316} by running several forward simulations and finding the best-fit. This is often slow and tough to scale to many parameters. Recent works have proposed new optimization methods for faster and more accurate calibration. For example, ~\cite{reiker2021machine} uses an ML model to identify what parameters to calibrate, however, values are identified with random search. ~\cite{vernon_bayesian_2022} use a gaussian process emulator which is computationally more efficient than the ABM. In a similar vein, ~\cite{abueg2021modeling} propose a surrogate strategy by first calibrating a compartmental SEIR-based model using standard ODE solvers, and supplying the resulting parameters to an ABM. However, such surrogate methods increase model miss-specification error as they cannot simulate stochasticity inherent in the ABM. Further, the SEIR-like models have few tunable parameters in comparison to ABMs, limiting the degree of calibration which can achieved. In contrast, the scalable and differentiable design of \ourmethod allows efficiently calibrating large-scale ABMs with gradient-based learning (and using deep neural networks).

\vspace{2mm}
\noindent\textbf{DNNs with ABMs.} Integrating deep neural networks (DNNs) with ABMs is an active direction. One popular approach has focused on multi-agent reinforcement learning (MARL), wherein the ABM is treated as a reinforcement learning environment, and a DNN is used to learn policies for the agents. Such perspective has been employed by ~\cite{ai_economist} to learn equitable economic policies, ~\cite{sert2020segregation} to analyze societal segregation dynamics, and ~\cite{oil_gas_macro} to learn oil and gas macro strategies. An alternate strategy has focused on using DNNs to emulate agent-based models ~\cite{bengio2020predicting, angione2022using, reiker2021machine} by using the ABM to generate training dataset for DNNs. Recently, ~\cite{bengio2020predicting} used an epidemiological ABM to generate synthetic datasets and trained DNN models to predict infectiousness for contact tracing. ~\cite{reiker2021machine} used a similar strategy (of DNN as surrogate) to identify what simulation parameters to calibrate. Both these directions (MARL, Surrogate) are constrained by non-differentiability of the ABM and hence cannot utilize automatic differentiation . In contrast, we introduce differentiable ABMs (\ourmethod) which allows joint training of DNNs with the ABM. This makes it possible to integrate heterogeneous data to infer latent micro variables and calibrate simulation parameters through hybrid DNN-ABM pipelines.

\subsection{Automatic Differentiation for Simulations}
Differentiable simulation is a powerful family of techniques that applies gradient-based methods to learning and control of physical systems. With progress in automatic differentiation, as computing gradients become easier, such simulators are emerging across wide range of systems. In combination (and joint training) with deep neural networks, such simulators are being used from computer graphics for looking through scattering media~\cite{diff_sim_vis1} and reconstructing human body pose~\cite{diff_sim_vis2}; to molecular dynamics for analysing the force field of ionic liquids~\cite{schoenholz2020jax, diff_sim_md2}; to robotics for grasping and locomotion~\cite{diff_sim_robotics, diff_sim_robotics2}. Even in epidemiology, gradient-based optimization with epidemiological simulators has also been recently explored with SIR-like models to learn simulation parameters~\cite{arik2020interpretable,qian2020and} and for making neural models to learn epidemic dynamics from mechanistic models~\cite{rodriguez_einns_2022}. \ourmethod aims to realize differentiable simulations for stochastic agent-based models. In contrast to relevant recent work in ~\cite{arya2022automatic, andelfinger2021differentiable}, \ourmethod obtains gradient estimates through stochasticity by efficiently reparameterizing with the gumbel-softmax gradient estimator (unlike smoothing in ~\cite{andelfinger2021differentiable} which has exponential cost scaling), shows scalability to million-size populations (unlike 25 agents in~\cite{arya2022automatic}) and also demonstrates ability to integrate with deep neural networks to calibrate large epidemiological ABMs.

\subsection{Invariances, Computation and ABMs}
As a result of leveraging the structure of the physical world, DNN architectures are effective in overcoming the curse of dimensionality~\cite{bronstein_geometric_2017}.
This is key to efficient computation on grids with CNNs (translation invariance) and graphs with GNNs via neural message passing (permutation invariance). We posit that while useful for learning DNNs, utilizing these invariances can make computation tractable in physical systems, where they exist naturally. Specifically, we observe that epidemiological models~\cite{deepabm, hinch2021openabm, abueg2021modeling, romero-brufau_public_2021} also adhere to permutation invariance wherein the order of infectious interactions (pair-wise message passing) within a step does not matter when estimating the probability of infection from those interactions. Motivated by this observation, we implement the transmission model of \ourmethod as a differentiable message passing operation defined with physical equations instead of DNNs

\section{Differentiable Agent-based Modeling for Epidemiology}
\begin{figure*}
    \centering
    \includegraphics[width=\textwidth]{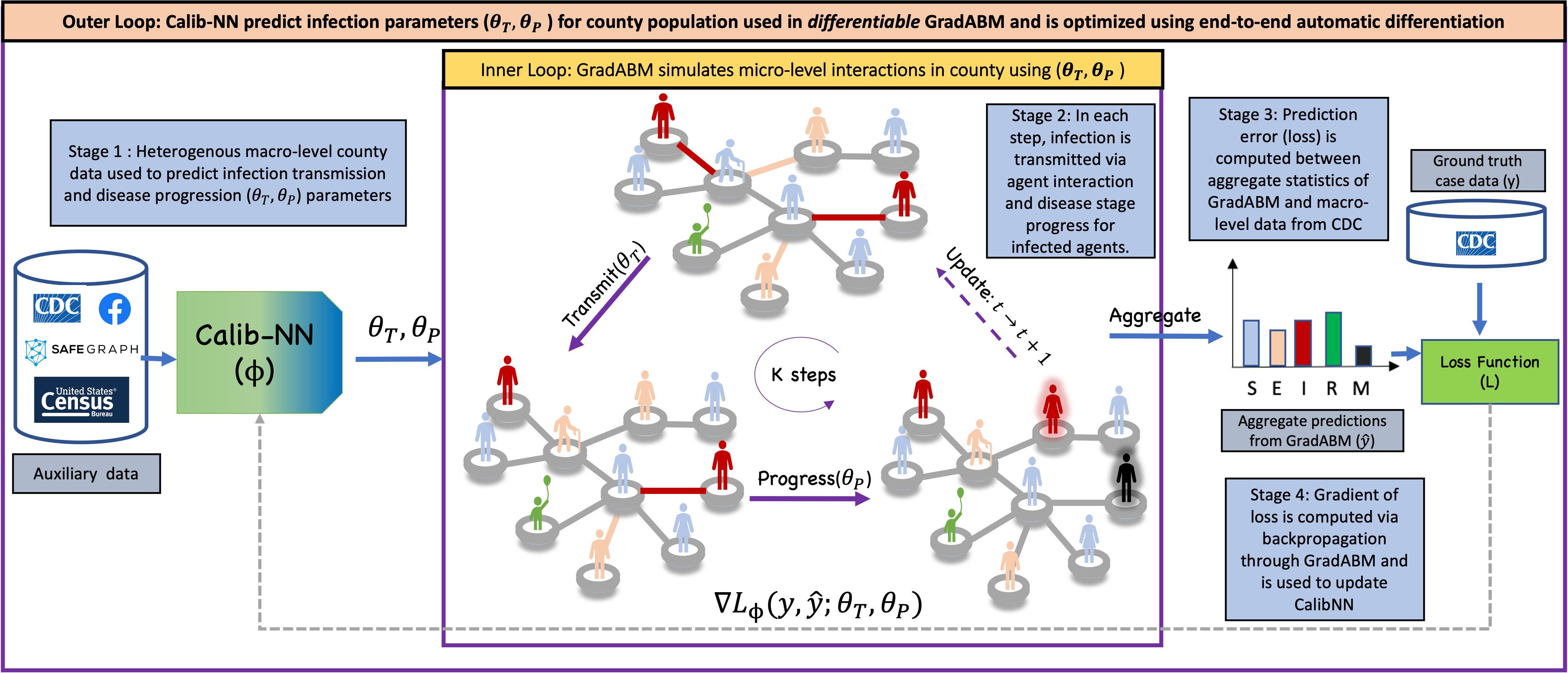}
    \caption{\textbf{Differentiable Agent-based Epidemiology} involves 4 stages i) heterogeneous macro-level population data (CDC, census, behavioral, survey) is input to a calibration model (\calibnn) to predict epidemiological parameters ($\theta_T, \theta_P$). ii): ($\theta_T, \theta_P$) are used run $K$ forward steps of the fully-differentiable epidemiological model (\ourmethod) which simulates micro-level infection transmission ($\texttt{Transmit}$) and disease progression ($\texttt{Progress}$) over individual contact networks. Disease statistics are aggregated ($\texttt{Aggregate}$) at end of $K$ steps to obtain the macro-level simulation output ($\hat{y}$). iii) Error between predicted $\hat{y}$ and real-world case statistics ($y$) is used to define a loss ($L(\hat{y}; \hat{y})$). iv) Gradient of this loss is computed by automatic differentiation through the micro-level \ourmethod to update weights of macro-level \ourmethod using gradient descent. }
    \label{fig:training_pipeline}
\end{figure*}

In what follows, we present a framework for epidemiological modeling with differentiable ABMs. The pipeline is summarized in Fig.~\ref{fig:training_pipeline}, where the inner loop is the differentiable epidemiological simulator (\ourmethod) and the outer loop is the gradient-based calibration procedure using the neural network \calibnn. The epidemiological simulator is described in Sec.~\ref{sec:epi-model}, the calibration procedure in Sec.~\ref{sec:calibnn} and the use of the final model for forecasting and evaluating policy interventions in Sec.~\ref{sec:forecast}.

\subsection{\ourmethod: Epidemiological Simulator} \label{sec:epi-model}
Here, we describe the inner loop shown in Fig.~\ref{fig:training_pipeline}. We consider a $K$-step discrete-event simulation with a population of $n$ interaction agents. The key contribution of this work is \ourmethod: an agent-based modeling design amenable for gradient-based learning with automatic differentiation. To achieve this, \ourmethod follows a tensorized implementation where each agent state is represented as a vector, the interaction networks as adjacency matrices and all discrete distributions (e.g., Bernoulli) are reparameterized with continuous approximations (Gumbel-Softmax~\cite{gumbel_softmax}), for end-to-end differentiability. As we discuss later, this design allows scaling agent-based models to realistic population sizes and integrating them with deep neural networks, which provides an array of practical benefits. Before proceeding, we first describe the epidemiological model used in our implementation which follows from ~\cite{deepabm, hinch2021openabm, romero-brufau_public_2021}

The state of each agent $j \in \{1, \dots,  n\}$, at time step $t \in \{0, \dots, K\}$, is denoted by a 3-D vector $X_{j}^{t} = \{ a_{j}, d_{j}^{t}, e^t_{j} \}$ where $a_{j}$ $\in \{0-10, 11-20, 21-30, \ldots, 71-80, 80+\}$ is the age; $d_{j}^{t}$ is the current disease stage; and $e^t_{j} \in \{-1, \dots, t-1\}$ denotes the time step of last exposure. For example, in COVID-19, $d_{j}^{t}$ can take values in $\{S, E, I, R, M\}$ which denote susceptible (S), exposed (E), infected (I), recovered (R) and mortality (M) stages. Agent state may update due to change in disease stage resulting from new infectious interaction with an exposed or infected agent (captured by the transmission function) or the natural progression of a previously incubated infection (captured by the progression function). This simulator is parameterized with time-dependent parameters which govern the transmission function ($\theta^t_T = [R^t, S_a, T_s, i^0]$) and progression function ($\theta^t_P = [\tau_{EI}, \tau_{IR}, \tau_{IM}, m]$). Here $R^t$ is the time-dependent disease reproduction number (measure of infection rate), $S_a$ is the age-stratified susceptibility, $T_s$ is the disease stage-stratified infection transmissibility, $i^0$ is the percentage of infections at time $t=0$, $\tau_{EI}$, $\tau_{IR}$, and $\tau_{IM}$ are the pairwise stage transition times from E to I, I to R and I to M, respectively, and $m$ is the mortality rate (fraction of expired agents amongst all agents in the $\{R,M\}$ disease stage). At each step $t$, every agent $i$ interacts with a set of neighbors $\{j: j \in \mathcal{N}_{i}\}$ from which she/he may accumulate or transmit infection. This disease stage evolution, at any step $t$, is given by:
\begin{equation}
    d^{t+1}_{i} = \texttt{Update}(X^{t}_{i}, \mathcal{N}_{i}, (X^{t}_{j})_{j \in \mathcal{N}_i}, \theta^t_T, \theta^t_P),
\end{equation}
where $\texttt{Update}(X^{t}_{i}, \mathcal{N}_{i}, (X^{t}_{j})_{j \in \mathcal{N}_i}, \theta^t_T, \theta^t_P)$ is
\begin{align}
 = 
    \begin{cases}
        \texttt{Transmit}(X^{t}_{i}, \mathcal{N}_{i}, (X^{t}_{j})_{j \in \mathcal{N}_{i}}, \theta^t_T), & \text{if } d^{t}_{i} = S,\\
        \texttt{Progress}(X^{t}_{i}, \theta^t_P), & \text{if } d^{t}_{i} \in \{E, I\}.
    \end{cases}
\end{align}

The transmission function $\texttt{Transmit}(X^{t}_{i}, \mathcal{N}_{i}, (X^{t}_{j})_{j \in \mathcal{N}_i}, \theta_T)$ computes the probability of infection transmission as a result of interactions between susceptible and infected (or exposed) agents. This probability  is proportional to an aggregate rate of transmission accumulated over the multiple interactions and is defined as 
\begin{equation}
    q(t) = 1 - e^{-\lambda_{A}}
\end{equation}
where $\lambda_{A} =  \bigcup_{j \in \mathcal{N}_i} \big( \lambda(R^t, S_i, T_j, e^t_{j} ) \big)$. Here, $\lambda$ denotes the rate of transmission from a single interaction and $\bigcup$ is an aggregation function which accumulates transmission over multiple interactions. In our implementation, $\lambda$ is a linear function as defined in ~\cite{abueg2021modeling} and $\bigcup$ is the summation ($\sum$) function following the invariance of infection transmission noted in Sec.~\ref{sec:rw} (more details in appendix). Before proceeding, the probability $q(t)$ is used to sample the agent's new disease stage from a Bernoulli random variable as $\hat{Q} = \mathrm{Bernoulli}(q(t), 1-q(t))$. Once infected, the agent can now infect other agents on subsequent steps and also enters a hierarchy of disease stage progression which triggers subsequent changes in the state of the agent. The progression function $\texttt{Progress}(X^{t}_{i}, \theta_P)$ updates the disease stage from $E \to I$ or from $I \to \{R, M\}$ at times determined by the stage transition time parameters, which may also be stochastic and require sampling from discrete distributions. This sequence of transmission and progression is repeated for $K$ steps of the simulation. Finally, the aggregate cumulative deaths is determined as: $\hat{y} = \texttt{Aggregate}(d^{T}_{i}) = m \times (d^T_i \in \{R,M\})_{i \in \{1, \dots, n\}}$, where $m \in \theta_P$. Ensuring differentiability of \ourmethod requires gradient estimation of discrete stochasticity. For this we replace the non-differentiable sample from the Bernoulli distribution with the Gumbel-Softmax gradient estimator~\cite{jang2017categorical, gumbel_estimator_review}.

\subsection{Training and Calibration of \ourmethod}
\label{sec:calibnn}
In order to use \ourmethod for real-world applications, we need to be able to calibrate its parameters ($\theta^t_T, \theta^t_P$) in such a way that the aggregate predicted quantities from the model such as cumulative deaths match the observed values from real-world data. Here, the differentiability of \ourmethod enables the possibility of utilizing gradient-based learning to calibrate the simulator by integrating with deep neural networks. Further, we note that the tensorized specification of \ourmethod is also critical since fast forward simulations are required to this iterative optimization where the simulator is executed at each optimization step. The calibration protocol is visualized in the outer loop in  Fig.~\ref{fig:training_pipeline} and described below. This outer training and calibration loop runs for $W$ optimization steps, with each step $w \in \{1,..,W\}$ comprising of the following four stages: 

\paragraph{Stage 1}: 
At the beginning of each outer loop, candidate parameters $(\theta^t_T, \theta^t_P)^w_{t \in \{0, \dots, K\}}$ are generated by a calibration neural network (\calibnn), whose weights at time step $w$ we denote by $\phi^w$.  To aid the generation of parameters, the calibration network is passed heterogeneous data as input. This may include data sources that were not considered during the initial construction of the underlying ABM being calibrated. For example, the calibration neural network may be passed social network data (e.g., self-reported symptom surveys conducted by Facebook) that was not integrated into the ABM, but that may still be of value when predicting disease prevalence. Intuitively, one may view this calibration network as a domain expert who integrates new data that the ABM designer was previously unaware of.
\paragraph{Stage 2}: 
After parameters have been generated, they are passed to \ourmethod in order to perform a $K$ step simulation, as described previously as the the inner loop in Fig.~\ref{fig:training_pipeline}. This outputs the aggregate predictions from the simulator which includes cumulative infections, deaths etc.
\paragraph{Stage 3}: 
The cumulative deaths across the $K$ steps of simulation are then compared with corresponding ground truth values. More specifically, a mean-squared loss between the simulated and actual cumulative deaths is computed. This is defined as: $\mathcal{L}(\hat{y}^w, y; (\theta^t_T, \theta^t_P)^w) = \texttt{MSE}(\hat{y}^w, y)$, where $y$ is the ground truth data representing cumulative deaths, $\hat{y}^w$ is the cumulative deaths predicted by \ourmethod at the $w^{\textup{th}}$ training step with input parameters $(\theta^t_T, \theta^t_P)^w)$ and $\texttt{{MSE}}$ denotes the mean-squared error function. 
\paragraph{Stage 4}: 
Since \ourmethod is differentiable (as established earlier), we may calculate the gradient of the computed loss with respect to each parameter of \calibnn, via backpropagation. Moreover, due to \ourmethod's network-centric tensorized design, auto-diff packages from libraries such as PyTorch can be leveraged to compute these gradients efficiently in a highly parallelized manner.
The final step of this gradient computation involves back-propagation through the weights of the \calibnn neural network ($\phi$) as well. 

Once, the gradient of the loss is computed, we use the classical gradient descent algorithm to update the weights of the calibration neural network at the end of each iteration as follows:
\begin{equation}
    \phi^{w+1} = \phi^{w} - \alpha \frac{\partial \mathcal{L}(\hat{y}^w, y; (\theta^t_T, \theta^t_P)^w)}{\partial \phi},
\end{equation}

where $\alpha$ is the learning rate, $(\theta^t_T, \theta^t_P)^w = f(D; \phi^w)$ and $\hat{y}^w$ is computed by calling $\texttt{Update}$ $K$ times with same parameters $(\theta^t_T, \theta^t_P)^w$. In essence, calibration here involves optimizing weights of \calibnn ($\phi$) which predicts the simulation parameters ($\theta_T, \theta_P$) instead of the parameters directly. We are essentially training the calibration neural network to understand heterogeneous data as a means of producing good parameterizations for \ourmethod. We call this calibration pipeline involving \calibnn as deep calibrated \ourmethod (DC-\ourmethod).

Consider for instance the scenario with multiple simulators in an experiment. This is realistic since a distinct \ourmethod may be defined for each county in a state. We now have the opportunity to jointly calibrate parameters for all counties using a shared \calibnn. Given $T$ counties, the weights of \calibnn $\phi$ may be trained on the average of the loss for all counties (Stage 2). The same weights of \calibnn can be used to predict personalized parameters for each county-specific \ourmethod. We call such a multi-task calibration procedure as joint deep calibrated \ourmethod (JDC-\ourmethod).

Note that each step of gradient descent requires a full forward simulation of \ourmethod. As a result, \ourmethod's network-centric design and tensorized implementation are key in ensuring that a sufficient number of gradient steps can be quickly computed to effectively train the calibration network. Additionally, observe that this optimization strategy is not possible with classical object-oriented ABMs, as they do not allow for scalable gradient computation.

\begin{remark}
We note that the \calibnn based approach presented in this paper is fundamentally distinct from emulation or surrogate models. Emulation models take the same input as the ABM's input and predict the output of the ABM, without actually simulating any agent behavior. In constrast, the output of \calibnn serves as the input to \ourmethod, which then simulates the behavior of the agents. Thus, \calibnn extends simulation pipeline, by enabling us to learn the correct inputs for \ourmethod.
\end{remark}

\subsection{Forecasting and policy evaluation} \label{sec:forecast}
Once trained the hybrid framework we propose, consisting of \calibnn and \ourmethod, it is ready to make forecasts in new, previously unseen scenarios as well as help decide on intervention policies.
These unseen scenarios may include predicting future the evolution of multiple infections for a county/region, 
or it may involve making predictions even when the data (inputs to \calibnn) for a county/region is noisy. In each of these scenarios, the data available for the county/region is fed to \calibnn, which then outputs personalized time-varying parameters $\theta^t_T$ and $\theta^t_P$ for $t\in\{0,...,K+H\}$, where $H$ is the forecasting horizon. These parameters are used in \ourmethod, which is run for $K+H$ steps, where the last $H$ simulation steps will serve as our forecasts.

\section{Experimental Setup}
\label{sec:setup}
To illustrate the effectiveness of our framework, we conduct experiments for COVID-19 and influenza on multiple counties of the state of Massachusetts, USA, learning personalized parameters for each county. 
Before presenting our analyses, we provide details regarding our experimental setup. 

\subsection{Heterogeneous Data Sources}
\label{subsec:data}
We describe the data sources passed as input to CalibNN, used to construct the population for \ourmethod and parameterize disease transmission and progression. (i) To start, we outline the features used as input to CalibNN during our experiments. In the case of COVID-19, CalibNN receives 5 input signals \footnote{Data links: \scriptsize \url{delphi.cmu.edu}; \url{goo.gle/covid19symptomdataset}; \url{safegraph.com}; \url{coronavirus.jhu.edu};\url{gis.cdc.gov/grasp/fluview/fluportaldashboard.html}} including insurance claims data, online symptoms surveys from Facebook, and line-list data.  Since all of these datasets contain granular information regarding each specific county, the data points input to CalibNN varies based on the county being modeled. Meanwhile, for influenza, CalibNN receives 14 signals originating from the Google symptoms dataset~\cite{google_symptoms}. This data is reported at the state level and hence we use the same data for all counties. (ii) In order to construct GradABM, several datasets are used. Demographic information from the US Census~\cite{us_census} is used to generate a synthetic agent population. Agent interaction graphs are created using census block level mobility data from Safegraph and leveraging demographic information (such as age) inside each census block. (iii) With respect to parameters in the transmission and progression models, we utilize clinical data from reliable sources such as the CDC~\cite{cdc_flu_facts} and clinical papers~\cite{hinch2021openabm,ferretti2020quantifying}. The target variable for COVID-19 is the COVID-associated mortality, whilst for influenza we use influenza-like-illness (ILI) counts collected by the CDC. Ground truth data for both target variables is obtained from the JHU and the CDC. 

\subsection{Baselines}
To showcase the performance of our calibration procedure (the outer loop described in Fig. ~\ref{fig:training_pipeline}), we compare \ourmethod with numerous benchmarks inspired by popular approaches for calibrating an ABM. \textbf{ExpertSearch-ABM:} Following~\cite{hinch2021openabm, romero-brufau_public_2021}, it uses parameters set by a combination of expert advice and randomized search. For COVID-19, we get $R_{0}$ and case-fatality rate from \cite{billah2020reproductive,abdollahi2020temporal} and $R_{0}$ for flu from \cite{chowell2008seasonal}. \textbf{SurrogateODE-ABM:} Following~\cite{abueg2021modeling}, it calibrates parameters by using a compartmental ODE model to emulate the ABM. We use SEIRM and SIRS models for COVID-19 and influenza respectively. Lastly, we compare three versions using \ourmethod that leverage gradients during the calibration process, namely calibrated \ourmethod (C-\ourmethod), deep calibrated \ourmethod (DC-\ourmethod) and jointly deep calibrated \ourmethod (JDC-\ourmethod) respectively. \textbf{C-\ourmethod:} directly optimizes the simulation parameters via gradient descent, without a calibration neural network (CalibNN). To clarify, the gradient update (in Stage 4) is given as: $[\theta^t_T, \theta^t_P]^{w+1} = [\theta^t_T, \theta^t_P]^w - \alpha \frac{\partial \mathcal{L}(\hat{y}^w, y; (\theta^t_T, \theta^t_P)^w)}{\partial [\theta^t_T, \theta^t_P]^w}$. \textbf{DC-\ourmethod:} calibrates the parameters by optimizing weights of CalibNN using automatic differentiation, as described in Sec.~\ref{sec:calibnn}. \textbf{JDC-\ourmethod:} jointly calibrates parameters for all counties using a shared CalibNN (multi-task learning), as in Sec.~\ref{sec:calibnn}. Here, instead of a new CalibNN for each county, a single CalibNN is shared. Intuitively, one may hope that such a calibration network learns to transfer useful information about disease spread from one county when reasoning about another.

\subsection{Metrics}
In order to provide a rigorous evaluation for the aforementioned calibration procedures, we use several standard metrics for evaluating epidemic predictions~\cite{tabataba_framework_2017,adhikari_epideep:_2019}. Specifically, we use normal deviation (ND), root mean squared error (RMSE) and mean absolute error (MAE). We provide further details in our appendix. Following CDC forecasting guidelines~\cite{biggerstaff_results_2016,cramer2022evaluation}, we make weekly predictions for 1 to 4 weeks ahead in the future. Our evaluation for both diseases is of at least 4 months in 10 counties. More details regarding the periods of evaluation and counties investigated are in the appendix. 

\subsection{Implementation Details}
In each of our experiments, we allow ABM parameters to change at each time step. That is, quantities such as the disease reproduction number and age-stratified susceptibilities are free to vary as the epidemic progresses. Consider the following intuitive argument for this design choice. It is reasonable to believe that individuals take greater precautions as an epidemic progresses. For example, many people began washing their hands more frequently during the COVID-19 pandemic, reducing the risk of disease transmission. Whilst \ourmethod does not explicitly model hand washing, parameters such as age-stratified susceptibility can be varied through time to account for its effect, especially if time series data related to sanitary practices is passed as input to CalibNN. Put differently, by allowing for ABM parameters to vary through time, we may account for environment dynamics or changes in agent behavior that aren't explicitly modeled by the simulator. We stress that our calibration framework does not depend upon CalibNN's neural architecture and that end-to-end training via a combination of backpropagation and gradient descent is possible regardless of CalibNN's implementation. In our own experiments, we design CalibNN so that it may take advantage of the heterogeneous time series data it is passed as input. More specifically, we employ an encoder-decoder architecture, based on GRUs and self-attention~\cite{vaswani2017attention}. More details regarding our choice of CalibNN architecture can be found in the appendix. \footnote{Our code and data is publicly available: \url{https://github.com/AdityaLab/GradABM}}

\section{Results}
\textit{First}, we show that calibrating with \calibnn can improve \ourmethod's forecasting accuracy significantly. Moreover, we show that \ourmethod scales efficiently as the number of interactions between agents grows. Next, we also show that our calibration procedure is robust to noisy data, in the sense that \ourmethod can be well calibrated even with observational error in ground truth; and that \ourmethod can be used to conduct sensitivity analyses for evaluating the sensitivity of policy interventions.

\begin{table*}[t]
\caption{Forecasting results for COVID-19 and influenza over 5 runs. JDC-\ourmethod is the only one consistently among the best performing for all  (lower error metrics is better).} 
\centering
\vspace{+.1cm}
\resizebox{0.95\linewidth}{!}{
\begin{tabular}{@{\extracolsep{4pt}}l|ccc|ccc}
\toprule   
{} & \multicolumn{3}{c|}{COVID-19}  & \multicolumn{3}{c}{Influenza}\\
 \cmidrule{2-4} 
 \cmidrule{5-7}  
 Model & ND & RMSE & MAE & ND & RMSE & MAE \\ 
\midrule
ExpertSearch-ABM ~\cite{romero-brufau_public_2021} & 8.75 & 689.92 & 270.13 & 0.57 & 2.03 & 1.72 \\
SurrogateODE-ABM ~\cite{abueg2021modeling}  & 2.21 $\pm$ 1.36   & 121.87 $\pm$ 63.97 & 68.20 $\pm$ 41.84 & 0.59 $\pm$ 0.02 & 2.17 $\pm$ 0.05 & 1.77 $\pm$ 0.05 \\
\midrule \midrule
\textbf{JDC-\ourmethod} & \textbf{0.97 $\pm$ 0.18}  & \textbf{50.99 $\pm$ 12.12} & \textbf{30.02 $\pm$ 5.60} & \textbf{0.41 $\pm$ 0.02} & \textbf{1.47 $\pm$ 0.06} & \textbf{1.22 $\pm$ 0.06} \\
\midrule
\textbf{DC-\ourmethod} & 1.15 $\pm$ 0.24 & 67.09 $\pm$ 23.89 & 35.50 $\pm$ 7.36 & 0.50 $\pm$ 0.19 & 1.78 $\pm$ 0.62 & 1.50 $\pm$ 0.57 \\
\textbf{C-\ourmethod} & 2.39 $\pm$ 0.35 & 205.14 $\pm$ 42.56 & 73.66 $\pm$ 10.88 & 0.88 $\pm$ 0.14 & 2.97 $\pm$ 0.44 & 2.64 $\pm$ 0.43 \\
\bottomrule
\end{tabular}
}
\vspace{-.05cm}
\label{tab:results}
\end{table*}

\subsection{Forecasting multiple infectious diseases}
Table~\ref{tab:results} displays the results of calibration for each of the benchmarks discussed in Section~\ref{sec:setup}. Observe that JDC-\ourmethod outperforms other benchmarks across all metrics for both Influenza and COVID-19 (rows 1-3 in Table~\ref{tab:results})\footnote{We performed the unpaired t-test test ($\alpha=0.05$) over 5 runs to verify JDC-\ourmethod performance gains w.r.t other ABM calibration methods are statistically significant}. For instance, JDC-\ourmethod improves ND to 0.97 from 8.75 in ExpertSearch-ABM and 2.21 in Emulator-ABM on COVID-19; and to 0.41 from 0.57 in ExpertSearch-ABM and 0.59 in Emulator-ABM on Influenza. More generally, we observe that JDC-\ourmethod achieves a consistent gain of 8x to 12x over ExpertSearch-ABM on all metrics in COVID-19 experiments. This performance benefit can be attributes to three key features which allows learning personalized time-varying parameters: gradient-based calibration, integration of heterogeneous data sources and joint county training via shared weights of \calibnn.

Meanwhile, by comparing JDC-\ourmethod to both DC-\ourmethod and C-\ourmethod (rows 3-5 in Table~\ref{tab:results}), we make some immediate observations. Firstly, using \calibnn in tandem with \ourmethod (JDC-\ourmethod, DC-\ourmethod) yields meaningful performance improvements over training \ourmethod directly via gradient descent (C-\ourmethod). Specifically, there is significant improvement in performance in DC-\ourmethod over C-\ourmethod across all metrics (eg: RMSE goes from 205.99 to 67.09). We conjecture that this is due to the ability of \calibnn's to alleviate overfitting by representing parameters in a high-dim space of DNN (instead of scalars in C-\ourmethod), inducing a joint distribution over all parameters and incorporating heterogeneous data sources when suggesting parameters. Moreover, we observe that joint training in JDC-\ourmethod leads to marginal improvement in performance over DC-\ourmethod (eg: RMSE goes from 67.09 to 50.99 in COVID-19 experiments). This indicates that \calibnn can implicitly use data from one county in order to improve calibration for others (through shared weights).

\subsection{Robustness, Scalability and Decision Making}
\label{sec:decisions}

\begin{figure}
    \centering
    \includegraphics[width=0.3\textwidth]{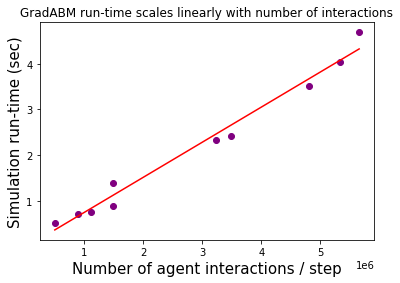}
  \caption{\ourmethod run-time scales linearly with the number of interactions and is roughly 300x faster than prior-art.  
  This benefit is due to the sparse-tensor calculus based design.}
  \label{fig:performance}
\end{figure}

Next, we study the practical applicability of our framework by focusing on three key considerations: i) scalability to large  populations, ii) robustness to observation error in data and iii) utility for practical decision making via sensitivity analyses of policy interventions.

\subsubsection{Scalability of \ourmethod simulations}
To investigate scalability, we examine how the simulation time of \ourmethod scales as the number of agent interactions increases. As is evident from Figure~\ref{fig:performance}, the run-time for \ourmethod scales linearly with number of interactions in the population and executes very quickly (even as the adjacency matrix scales quadratically). For instance, \ourmethod executes a simulation with 800,000 agents (5 million interactions) over 133 steps in 4 seconds on a GPU (and 60 seconds on CPU). This is roughly 300x faster than the equivalent Mesa~\cite{mesa} implementation. This performance improvement can be attributed to \ourmethod's sparse tensor-calculus implementation. Since forward simulation is efficient, sensitivity analysis of GRADABM can be conducted rapidly via repeated simulations with different parameter settings. Moreover, iterative algorithms which perform a forward simulation at each step, such as gradient descent, can be used for calibration (as is demonstrated in our work).

\subsubsection{\ourmethod robustness to observational error}
\begin{figure}
\centering
             \includegraphics[width=0.35\textwidth]{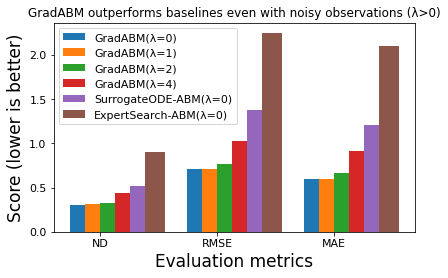}
        \caption{\ourmethod is robust to noisy data. \ourmethod achieves lower forecasting error than all baselines even when it is trained with noisy data ($\lambda > 0$) while the baselines receive original data. This is achieved due to differentiable design which allows encoding simulation parameters with \calibnn and learning using heterogeneous data sources.}
        \vspace{-.1cm}
         \label{fig:robustness}
\end{figure}

To investigate the robustness of our proposed calibration procedure, we run experiments using ground truth data distorted by gaussian noise. More specifically, we add gaussian noise to each ground truth target with mean $\mu=0$ and varying scales of standard deviation $s$. To set the standard deviation of noise for each county, we first compute the standard deviation of the ground truth data and multiply it by a $\lambda$ factor. In our experiments, we test four different values for $\lambda$. Results are presented in Figure~\ref{fig:robustness}. Even for a large degree of noise $(\lambda=4)$, we observe that JDC-ABM outperforms both Emulator-ABM and ExpertSearch-ABM on noiseless data. We attribute this to \calibnn. More specifically, we conjecture that \calibnn alleviates overfitting by representing parameters in high-dim space of the neural network (instead of scalars) and allows for integration of heterogeneous data that may assist in selecting appropriate parameters even when the ground truth is noisy.

\subsubsection{Evaluating sensitivity of policy interventions with \ourmethod}
Early in the COVID-19 pandemic, uncertainty in vaccine supplies forced policy makers to ask this question: \textit{what-if we delay administration of second dose of the COVID-19 vaccine?}. This decision is sensitive to several variables, including the protection offered by the first vaccine dose, which was unknown at the time~\cite{romero-brufau_public_2021, majeed2022implementation, hogan2020report}. Here, we elucidate how \ourmethod can help with decision making under uncertainty by enabling analysis of the sensitivity of policy interventions. More specifically, we consider a scenario with two alternative policies - P1: second dose is administered under standard schedule and P2: second dose is administered with a delay (more details in the appendix). Following the exact experimental setup from clinical works on this question~\cite{romero-brufau_public_2021}, we vary the protection of the first COVID-19 dose from 50\% to 80\% in ten percentile increments and run simulations for each of the four configurations under both policies. Then, we compare both policies by computing the ratio of cumulative deaths of P2 by P1, which we denote as relative mortality. Basically, if the relative mortality is less that 1, then policy P2 is better (can delay the second COVID-19 dose); while relative mortality greater than 1 implies that policy P1 is better (don't delay the second COVID-19 dose). Results for Franklin County, MA are shown in the appendix with additional details. We observe that once the protection of the first COVID-19 vaccine dose is greater than 60\%, \ourmethod recommends policy P2 (to delay the second dose). These results are consistent with past clinical works that evaluated this policy intervention~\cite{romero-brufau_public_2021,imai2021quantifying}.

\section{Conclusion and Future Work}
We introduced \ourmethod, a design for agent-based modeling that is amenable to gradient-based learning with automatic differentiation. \ourmethod achieves this via a tensorized implementation where each agent state is represented as a vector, the interaction networks as adjacency matrices and all discrete distributions (e.g., Bernoulli) are reparameterized for end-to-end differentiability. Experiments demonstrate that \ourmethod can quickly simulate realistic population sizes, integrate with deep neural networks and ingest heterogeneous data sources. This provides an array of practical benefits for calibration, forecasting, and evaluating policy interventions. 

\begin{figure}
    \centering
    \includegraphics[width=0.7\columnwidth]{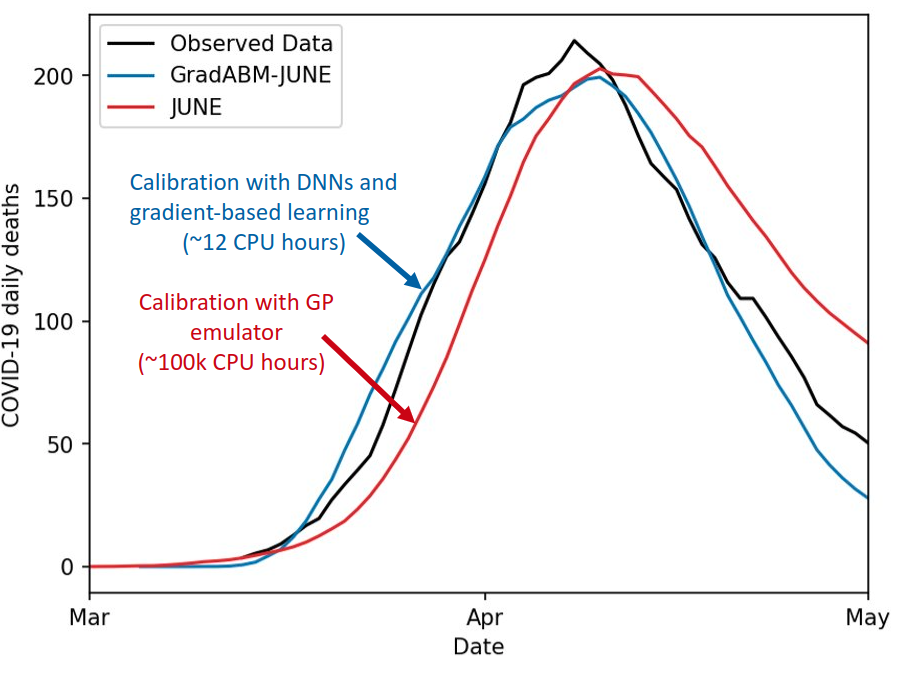}
    \caption{Daily deaths in London during the first wave of COVID-19 for the calibrated models JUNE~\cite{aylett2021june} and \ourmethod-JUNE  and observed data. Note we obtain the similar quality of fit with much less computing time.
    }
    \label{fig:june_london_fit}
\end{figure}

The compatibility of \ourmethod with automatic differentiation opens up multiple directions of future work both in terms of utility to epidemiology and for advancing the science of agent-based modeling. First, in our work, disease progression in an individual was modeled using a linear deterministic model, but future work could explore more complex and stochastic models (e.g., \cite{romero-brufau_public_2021}). Future work could also explore how to use gradient-based explainability methods~\cite{selvaraju2017grad} to have a better understanding of the underlying drivers of epidemic predictions. In this work, calibration with \calibnn predicts the maximum likelihood estimates but the our design allows us to also estimate the posterior distribution of the parameters using techniques from generative DNNs such as normalizing flows~\cite{normalizing_flows}. In addition, ABMs may be mis-specified for the actual disease dynamics for which we may have to change the model specification. In this work, we focused on \ourmethod using the OpenABM epidemiological model ~\cite{hinch_openabm-covid19agent-based_2021} designed for the US. As we explain next, \ourmethod generalizes to other model specifications. 

The JUNE epidemiological model~\cite{aylett2021june} is a large-scale agent-based model for epidemiology designed to support the National Health Service in England. It was also adopted by the UN Global Pulse to study refugee settlements ~\cite{aylett-bullock_operational_2021}. In the England setting, JUNE simulates the movement and interactions of ~55 million agents and features significantly more complex contact graphs than OpenABM, since it is based on the highest resolution of English census data. We were able to successfully translate it to the fully tensorized design by converting the contact structure in a heterogeneous graph by representing agents and locations as different node types. In this new implementation, which we refer to as \ourmethod-JUNE, we achieve a ~10,000x speed increase in simulation time. Due to the high computational cost of JUNE, the original model was calibrated using a surrogate model in the form of a Gaussian Process (GP) emulator ~\cite{vernon_bayesian_2022}. The differentiability of the new implementation coupled with the \calibnn pipeline allows us to quickly find parameter sets (reducing calibration time from 100,000 CPU hours to 12 CPU hours) that match the observed data well (Fig.~\ref{fig:june_london_fit}) and are comparable in quality to the fit of the GP emulator. This case study is an encouraging evidence on the utility of \ourmethod for creating real impact and also is a step towards our vision of making \ourmethod a domain-agnostic design for scalable and differentiable agent-based modeling.

\section*{Acknowledgements}
This work was supported in part by the NSF (Expeditions CCF-1918770, CAREER IIS-2028586, RAPID IIS-2027862, Medium IIS-1955883, Medium IIS-2106961, CCF-2115126, PIPP CCF-2200269), CDC MInD program, faculty research award from Facebook and funds/computing resources from Georgia Tech and MIT. AC was supported by Adobe Data Science Research Award. AQ was supported by a UKRI AI World Leading Researcher Fellowship awarded to Wooldridge (grant EP/W002949/1).
\small

\balance


\bibliographystyle{ACM-Reference-Format} 

\normalsize

\clearpage

\appendix
\normalsize

\section{Extended related work}

\par\noindent
\underline{\textbf{Permutation invariance: key insights and assumptions}}
We begin by highlighting some key assumptions that are followed by epidemiologists and then leverage them to motivate the specific implementation of \ourmethod. The key insight of \ourmethod is to utilize differentiable sparse-tensor computation in interaction networks and symmetries of the transmission model for fast and end-to-end differentiable simulations. Furthermore, we note that: i) clinical data shows that the natural infection takes multiple days to incubate, ii) CDC policy decisions are guided by 4-week ahead forecasting and analysis. These considerations motivate the following standard assumptions in epidemiological simulation work: \textbf{Assumptions}: Several works in epidemiological ABMs~\cite{deepabm, hinch2021openabm, abueg2021modeling, romero-brufau_public_2021} ratify the following assumptions: \textbf{A1}: the granularity of each step of the simulator is one day, \textbf{A2}: infectious interactions "do not collide" i.e. on any given day, an agent can not also infect other agents the day it is exposed, \textbf{A3}: an agent cannot be reinfected while already infected, \textbf{A4}: an agent cannot be infected when it is in the $RM$, i.e. recovered or mortality stage, i.e., there are no reinfections (this assumption can be relaxed based on the nature of the disease being modelled) \textbf{A5}: cumulative transmission over multiple interactions can cause infection, \textbf{A6}: infection does not accumulate over days.
It is important to note that real-world interaction networks routinely have low degree and high centrality and are modeled as small-world graphs as in~\cite{abueg2021modeling, romero-brufau_public_2021}. This allows use of sparse tensors to represent and process interaction networks; and scale to realistic populations (larger $n$) which improves performance when outputs are scale dependent. Also,  Assumptions \textbf{A2-A5} together imply that the order of interactions within a step does not matter. This enables us to invoke permutation invariance in interactions and makes the transmission model differentiable.

\section{End-to-end differentiability}
In practice, auto-diff packages such as Pytorch implement back-propagation through such a `for loop' by unrolling this loop in the internal computational graph that the library tracks to enable end-to-end differentiation. The only requirement is that all steps within the loop be differentiable, which we establish by showing that the components of our epidemiological model - Transmission model and Progression model are differentiable. 
The Transmission Model is differentiable as the functions comprising it, which are: $\lambda$ is a smooth function, $\bigcup$ is a permutation invariant function ($\sum$) and is linear. In order to make the sampling from $\hat{Q}$ differentiable, we utilize a continuous relaxation with Gumbel-Softmax reparametrization~\cite{gumbel_softmax}. Next, the Progression Model is differentiable as it is effectively a linear deterministic model, with the parameters being the transition times. Thus, the entire $K$ step epidemiological model is differentiable.

\section{\calibnn details} 
For \calibnn, we encode the feature time series until $t_N$ by passing it through a Gated Recurrent Unit (GRU)~\cite{cho2014learning} to obtain a condensed representation for each time step:
\begin{equation}
    \{\textbf{h}^t\}_{t=t_0}^{t_N} =  \text{GRU}(\left\{\textbf{x}^{t}\right\}_{t=t_0}^{t_N} )
\end{equation}
where $\textbf{h}_t$ is the hidden state of the GRU for time step $t$.
To capture long-term relations and prevent over-emphasis on last terms of sequence we use self-attention layer:
\begin{equation}
    \left\{ \lambda^{t} \right\}_{t=t_0}^{t_N}  = \text{Self-Atten}(\{\textbf{h}^t\}_{t=t_0}^{t_N}), 
\label{eqn:atten}
\end{equation}
where Self-Atten~\cite{vaswani2017attention} involves passing the embeddings into linear layers to extract meaningful similarities before normalizing the similarities using Softmax.
Then, we use the attention weights to combine the latent representations and obtain a single embedding representing the time series of features from $t_0$ to $t_N$: $\textbf{u}^{t_0:t_N} = \sum_{t=t_0}^{t_N}\lambda_t \textbf{h}_t,$
\begin{gather}
\label{eq:attn}
\textbf{U} =  \text{Softmax} \Big(\frac{\textbf{Q}      \textbf{K}^T}{\sqrt d_k} \Big) \textbf{V} \qquad \textbf{u}_{t_0:t_N} = \sum_{t=t_0}^{t_N} \textbf{u}^{t}
\end{gather}
therefore, our vector $\textbf{u}^{t_0:t_N}$ summarizes the input sequence and represents the context to be given to the decoder. Next, we use another GRU as a decoder, which takes the context $\textbf{u}^{t_0:t_N}$ and a positional encoding $\tau_k$ that informs the model of how many days in the future we want to predict. We simply use $\tau_k$ to be a float between 0 and 1. 
\begin{equation}
    \{\textbf{o}^t\}_{t=t_1}^{t_K} =  \text{GRU}(\{\tau^k\}_{k=1}^K; \textbf{u}^{t_0:t_N})
\end{equation}
We want to unroll our decoder to predict not only the values of the simulator, but also the future ones. For this, we pass the decoder output through a feedforward network $\text{FFN}(\textbf{o}^t)$.
We found the optimization can be challenging if we directly use the output of the neural network as parameters of our simulator. Therefore, we bound the output in a similar manner as \cite{arik2020interpretable}:
$\theta^t = \theta_L + (\theta_U - \theta_L) \cdot \sigma(\text{FFN}(\textbf{o}^t))$, where $\theta_L$ and $\theta_U$ are the lower and upper bounds of $\theta^t$ for all $t$, and $\sigma$ is the Sigmoid function.

\section{More details on experimental setup}
As noted in our paper, code and data are attached to this supplement and will be made publicly available upon acceptance. 

\par\noindent\textbf{Computational setup.}
All experiments were conducted using a 4 Xeon E7-4850 CPU with 512GB of 1066 Mhz main memory and 4 GPUs Tesla V100 DGXS 32GB. Our method implemented in PyTorch trains on GPU in ~15 mins. Inference takes only a few seconds.

\par\noindent\textbf{Real-time forecasting.}
We follow the literature on evaluating epidemic forecasting methodologies~\cite{shaman_real-time_2013,kamarthi_camul_2021,adhikari_epideep:_2019} and use the \emph{real-time forecasting} setup. 
We simulate real-time forecasting by making models train \emph{only} using data available until each of the prediction weeks and make predictions for 1 to 4 weeks ahead in the future. Data revisions in public health data are large and may affect evaluation and conclusions~\cite{kamarthi2021back2future,cramer2022evaluation}, therefore, we utilize fully revised data following previous papers on methodological advances~\cite{adhikari_epideep:_2019,rodriguez_steering_2021}.

\par\noindent\textbf{Evaluation.}
As evaluation is on weekly predictions, we aggregate daily predictions to obtain weekly predictions (sum for COVID-19 and average for flu). As stated in the main paper, we opted to use the same set of ABM parameters for every 7 days (i.e., we learn a new set of parameters for every week). This is a reasonable assumption as it has been often found that the parameters of mechanistic models do not change much from one day to the other~\cite{rodriguez_einns_2022}.
In our evaluation, we work with the following counties in Massachusetts: 25001, 25003, 25005, 25009, 25011, 25013, 25015, 25021, 25023, 25027. The specific evaluation period is determined with epidemic weeks\footnote{\url{https://ndc.services.cdc.gov/wp-content/uploads/MMWR_Week_overview.pdf}} which is the standard in CDC's epidemic prediction initiatives\footnote{\url{https://predict.cdc.gov/}}. 
For COVID-19 these are 202014, 202016, 202018, 202020, 202022, 202024, 202026, 202028, 202030. For flu, we evaluate in epidemic weeks 201746, 201748, 201750, 201752, 201802, 201804, 201806, 201808, 201810.
As noted before, this means that for each epidemic week, we make 4 predictions in the future.

\begin{table*}[h!]
\label{tab:strategies}
\small \centering
\caption{Visualization of different vaccination prioritization strategies we compare using \ourmethod}
\begin{tabular}{|l|l|l|}
\hline
Name of Strategy                                                                   & Description                                                                                                                                                                         & \begin{tabular}[c]{@{}l@{}}Priority list under \\ the strategy will be\end{tabular}    \\ \hline \hline
\begin{tabular}[c]{@{}l@{}}Standard dosing \\ (two doses on schedule)\end{tabular} & \begin{tabular}[c]{@{}l@{}}Age prioritized vaccination with second dose \\ at 21 days\end{tabular}                                                                                  & \begin{tabular}[c]{@{}l@{}}Betty, David, Frank, \\ Adam, Charlie, Eleanor\end{tabular} \\ \hline
Delayed second dose                                                                & \begin{tabular}[c]{@{}l@{}}Age prioritized vaccination prioritizing \\ vaccination of first-dose eligible individuals\end{tabular}                                                  & \begin{tabular}[c]{@{}l@{}}Adam, Charlie, Eleanor, \\ Betty, David, Frank\end{tabular} \\ \hline
\end{tabular}
\label{tab:vaccine-priority-graphic}
\end{table*}

\par\noindent\textbf{Metrics.}
Let $\hat{y}_{w,\tau}$ be the prediction week $w$ for $\tau$-weeks ahead in the future, $y_{w,\tau}$ be the corresponding ground truth value, and $e_{w,\tau} = \hat{y}_{w,\tau} - y_{w,k}$. In this paper $\tau$ takes from 1 to $T=4$. Then, we define our metrics as follows: 
$\text{ND} = \sum|e_{w,\tau}| / \sum|y_{w,\tau}|$, $\text{RMSE} = \sum e_{w,\tau}^2 / WT$ and $\text{MAE} = \sum|e_{w,\tau}| / WT$ .

\par\noindent\textbf{Implementation details.}
$\bullet$ \emph{Data preparation}:
All our time series data that enter to \calibnn are padded with a minimum sequence length of 20 for COVID and 5 for influenza. We also normalize each features with mean 0 and variance 1. To inform \calibnn of the county to predict, we use a one hot encoding for counties.
$\bullet$ \emph{Architecture details}: In \calibnn, the encoder is a 2-layer bidirectional GRU and decoder is a 1-layer bidirectional GRU, both with hidden size of 32. Output layer has two linear layers of size 32x16x$D$ with ReLU activation function and $D$ is ABM parameter dimensions dimensions: $D=3$ for COVID-19 and $D=2$ for flu. 
$\bullet$ \emph{Hyperparameters}: We found a learning rate of $10^{-3}$, Xavier initialization, and the Adam optimization algorithm work best. 
$\bullet$ \emph{ABM parameters $(\theta_P, \theta_T)$}: For COVID-19, we have three parameters: $R_0$, mortality rate, and initial infections percentage. These are bounded with $\theta_L=[ 1.0, 0.001, 0.01 ]$ and $\theta_U=[ 8.0, 0.02, 1.0 ]$. For flu, we have two parameters: $R_0$ and initial infections percentage. These are bounded with $\theta_L=[ 1.05, 0.1 ]$ and $\theta_U=[ 2.6, 5.0 ]$. The initial infections percentage is the percentage of the population that is infected at time step $t=0$ of the simulation.
$\bullet$ \emph{ABM clinical parameters}: 
To set some of the parameters of our transmission and progression models, we utilize clinical data from reliable sources. Specifically, for COVID-19 we use age-stratified susceptibility and parameters of a scaled gamma distribution to represent infectiousness as a function of time as per~\cite{hinch2021openabm,ferretti2020quantifying}. For influenza, we set those parameters based on CDC flu facts~\cite{cdc_flu_facts}.

\par\noindent\textbf{Target variables.}
The target variable for COVID-19 is COVID-associated mortality, while in flu we use influenza-like-illness (ILI) counts, which is collected by the CDC. ILI measures the percentage of healthcare seekers who exhibit influenza-like-illness symptoms, defined as "fever (temperature of 100°F/37.8°C or greater) and a cough and/or a sore throat without a known cause other than influenza"~\cite{cdc_flu_surveillance}.
The ground truth data for the target variables is obtained from JHU CSSE COVID-19 data repository and ILI from CDC influenza dashboard.

\par\noindent\textbf{Details on baseline implementation.}
\textbf{ExpertSearch-ABM:}
As noted in the main paper, $R_0$ and case-fatality rate are obtained from authoritative sources. 
To set the initial infections percentage for this baseline, we set it to the mean value of the search range.
\textbf{SurrogateODE-ABM: } We present details on each of the ODE models we used for this baseline. 

(COVID-19) SEIRM~\cite{wu2020nowcasting}: The SEIRM model consists of five compartments: Susceptible ($S$), Exposed ($E$), Infected ($I$), Recovered ($R$), and Mortality ($M$). It is parameterized by four variables $ \Omega = \{ \beta, \alpha,  \gamma, \mu\}$, where $\beta$ is the infectivity rate, $1/\alpha$ is the mean latent period for the disease,
$1/\gamma$ is the mean infectious period,
and $\mu$ is the mortality rate. The basic reproductive number  $R_0 = \beta/(\gamma+\mu)$.
\begin{gather}
\label{eq:seirm}
\frac{dS_t}{dt} = - \beta_t \frac{S_t I_t}{N} \qquad \frac{dE}{dt}  = \beta_t \frac{S_t I_t}{N} - \alpha_t E_t \\\nonumber 
\frac{dI_t}{dt} = \alpha_t E_t  - \gamma_t I_t - \mu_t I_t \qquad \frac{dR_t}{dt} = \gamma_t I_t \qquad \frac{dM_t}{dt} = \mu_t I_t
\end{gather}

(Influenza) SIRS~\cite{shaman2010absolute}: This model consists of three compartments: Susceptible ($S_t$), Infected ($I_t$), and Recovered ($R_t$). It is parameterized by three variables $ \Omega = \{ \beta, D, L\}$, where $\beta$ is the infectivity rate, $D$ is the mean duration of immunity, and $L$ is the mean duration of the immunity period. The basic reproductive number  $R_0 = \beta D$.
\begin{gather}
\frac{d S_t}{d t}=\frac{N-S_t-I_t}{L_t}-\frac{\beta_t I_t S_t}{N} \\\nonumber
\frac{d I_t}{d t}=\frac{\beta_t I_t S_t}{N}-\frac{I_t}{D_t}
\end{gather}

\section{Mode details on datasets}
The detailed description of datasets is as follows. 
\begin{itemize}
    \item Data signals 1: Mobility signals. The signals originate from the record of people visiting points of interest (POIs) in various regions. According to Google, daily changes in visits to various POI categories are collected and compared with the period January 3-February 6, 2020. Additionally, we collected a daily change of visitors from Apple, which shows the relative volume of directions requested across different US states compared to January 13. Different non-pharmaceutical interventions (NPIs) and different policies adopted by different states are implicitly illustrated by mobility signals.
    \item Data signals 2: Symptomatic surveys. Every day, Facebook collects statistics on COVID-like illness (CLI\%) and influenza-like illness (ILI\%) across the US and different states. On the basis of symptoms reported in voluntary surveys, they estimate this percentage.
    \item Data signals 3: Symptom search data. Google collects records of searches related to symptoms for multiple conditions and syndromes across the US and different states. Their system provides a metric that quantifies search volume associated with specific symptoms, which undergoes a privacy-protecting mechanism before being publicized. There are 400+ symptoms available in this dataset dating back to 2017, from which we only use a subset of 14. These are the following which are symptoms associated with influenza: Fever, Low-grade fever, Cough, Sore throat, Headache, Fatigue, Vomiting, Diarrhea, Shortness of breath, Chest pain, Dizziness, Confusion, Generalized tonic–clonic seizure, and Weakness.
    \item Data signals 4: Number of hospitalizations. The US Department of Health \& Human Services provides daily hospitalization admissions dating back to January 1, 2020. Several primary sources provide facility-level granularity reports to create this signal: (1) HHS TeleTracking, (2) reporting provided to HHS Protect by state/territorial health departments on behalf of their healthcare facilities, and (3) the National Healthcare Safety Network.
    \item Data signals 5: Number of new deaths. The Johns Hopkins University reports daily mortality for COVID-19. They collect and curate data from official websites of state public health departments across the US. This has been the source of data for the CDC COVID-19 forecasting initiative~\cite{cramer2022evaluation}.
    \item Data signals 6: weighted Influenza-like Illness (wILI). Time series data are collected by CDC from over 3,500 outpatient healthcare providers in the Outpatient Influenza-like Illness Surveillance Network (ILINet). Health care providers report voluntarily every week the percentage of patients with Influenza-like Illness (ILI) symptoms. ILI is defined as “fever (temperature of 100◦F [37.8◦C] or greater) and a cough and/or a sore throat without a known cause other than influenza.” This has been the source of data for previous iterations of the CDC FluSight forecasting initiative~\cite{reich2019collaborative}.
\end{itemize}

\section{More details on evaluating policy interventions}
We reproduce the experimental setup in~\cite{romero-brufau_public_2021}. For interventions, we simulate standard COVID-19 vaccination versus delayed second dose vaccination prioritizing the first dose. Sensitivity analyses included first dose vaccine efficacy of 50\%, 60\%, 70\%, 80\%, and 90\% after day 12 post-vaccination with a vaccination rate of 0.3\% population per day; assuming the vaccine prevents only symptoms but not asymptomatic spread (that is, non-sterilizing vaccine). We measure cumulative COVID-19 mortality, cumulative SARS-CoV-2 infections, and cumulative hospital admissions due to COVID-19 over 74 days. We explicitly modeled the confirmation of infections with polymerase chain reaction testing and quarantining of known infected agents with imperfect compliance over time. To simulate a natural pattern of infection at the point vaccinations begin, we started our simulation with 10 agents infected and ran the simulation for 20 days before starting vaccinations, which corresponds to a cumulative infection rate of 1\%, similar to the one in the US, UK, and most of Europe when vaccinations were started and corresponding to the time horizon used in our analysis. In both our vaccination strategies, we started administering vaccines on the basis of age, starting with people over 75, then those over 65, and so on.

We compare different vaccine regimens and prioritization schedules. Consider the following six hypothetical individuals with their age and dose eligibility on any given step of the simulations: i) Adam - first dose eligible 78 yr old; ii) Betty - second dose eligible 78 yr old; iii) Charlie - first dose eligible 68 year old; iv) David - second dose eligible 68 year old; v) Eleanor - first dose eligible 40 year old; vi) Frank - second dose eligible 40 year old. Table ~\ref{tab:vaccine-priority-graphic} shows the order in which these individuals are prioritized for vaccine administration under the strategies we compare.

\end{document}